\pdfoutput=1

\documentclass[11pt]{article}

\usepackage{acl}

\usepackage{times}
\usepackage{latexsym}

\usepackage[T1]{fontenc}

\usepackage[utf8]{inputenc}

\usepackage{microtype}

\usepackage{multirow}
\usepackage{booktabs}
\usepackage{pgfplots}
\usepackage{pgfplotstable}
\usepackage{tikz}
\usetikzlibrary{positioning}
\usetikzlibrary{backgrounds}
\usepackage{tikzscale}
\usepgfplotslibrary{groupplots}
\pgfplotsset{compat=newest}
\usepackage{amsmath,amssymb,amsfonts}
\usepackage{subcaption}
\usepgfplotslibrary{groupplots}
\usepackage{enumitem}
\usepackage{bm}
\usepackage{cleveref}
\crefformat{section}{\S#2#1#3}
\crefformat{subsection}{\S#2#1#3}
\crefformat{subsubsection}{\S#2#1#3}
\usepackage{xcolor}

\title{Cryptocurrency Bubble Detection: A New Stock Market Dataset, Financial Task \& Hyperbolic Models}

\author{Ramit Sawhney$^\dagger$\thanks{~~Equal contribution.}, Shivam Agarwal$^\ddagger$\footnotemark[1], Vivek Mittal$^\dagger$\footnotemark[1], \\
\textbf{Paolo Rosso$^\bigtriangleup$, Vikram Nanda$^\bigstar$, Sudheer Chava$^\dagger$} \\
$^\dagger$ Financial Services Innovation Lab, Georgia Institute of Technology,\\ $^\ddagger$University of Illinois at Urbana-Champaign, 
$^\bigtriangleup$Universitat Politècnica de València,\\ $^\bigstar$University of Texas at Dallas\\
  \texttt{shivama2@illinois.edu, \{rsawhney31,schava6\}@gatech.edu} \\}

\begin{document}
\maketitle
\begin{abstract}
The rapid spread of information over social media influences quantitative trading and investments. 
The growing popularity of speculative trading of highly volatile assets such as cryptocurrencies and meme stocks presents a fresh challenge in the financial realm. 
Investigating such "bubbles" - periods of sudden anomalous behavior of markets are critical in better understanding investor behavior and market dynamics.
However, high volatility coupled with massive volumes of chaotic social media texts, especially for underexplored assets like cryptocoins pose a challenge to existing methods. 
Taking the first step towards NLP for cryptocoins, we present and publicly release CryptoBubbles, a novel multi-span identification task for bubble detection, and a dataset of more than 400 cryptocoins from 9 exchanges over five years spanning over two million tweets.
Further, we develop a set of sequence-to-sequence hyperbolic models suited to this multi-span identification task based on the power-law dynamics of cryptocurrencies and user behavior on social media.
We further test the effectiveness of our models under zero-shot settings on a test set of Reddit posts pertaining to 29 ``meme stocks'', which see an increase in trade volume due to social media hype. 
Through quantitative, qualitative, and zero-shot analyses on Reddit and Twitter spanning cryptocoins and meme-stocks, we show the practical applicability of CryptoBubbles and hyperbolic models.
\end{abstract}

\section{Introduction}
\begin{figure}[t]
    \centering
    \includegraphics[width=\linewidth]{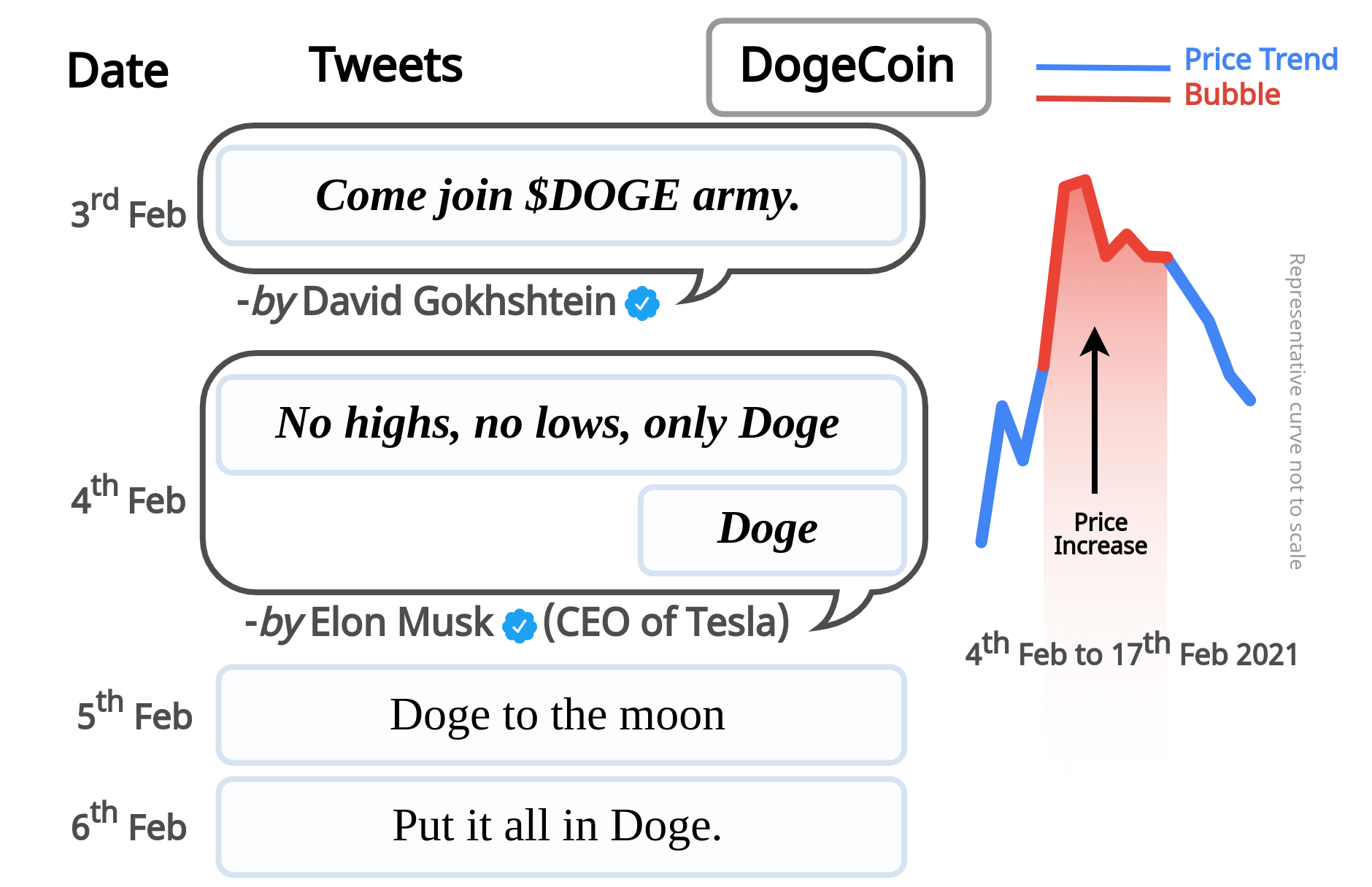}
    \caption{We present market moving tweets by influential users about DogeCoin. Such tweets induce social media hype which leads to creation of bubbles.}
    \label{fig:intro}
    \end{figure}

Cryptocurrency (crypto) trading presents a new investment opportunity \cite{chuen2017cryptocurrency} for maximizing profits. 
The rising ubiquity of speculative trading of cryptocurrencies over social media has lead to sentiment driven ``bubbles'' \cite{chohan2021counter,hu2021rise}. 
Such a bubble is characterized by rapid escalation of price in a  short period of time typically driven by exuberant investor behavior \cite{fry2016negative} and may be tied with enormous risks. 
Analyzing such anomalous behaviors can be useful for forecasting speculative risks.
However, it is not trivial to use existing NLP methods \cite{mansf} for forecasting crypto bubbles as they are designed for simple assets such as equities that are 10x less volatile than crypto \cite{Liu2019}.
Also, crypto behavior is more strongly tied to user sentiment and social media usage as opposed to conventional stocks and equities, rendering both conventional financial models and contemporary ML models weak as they are not geared towards dealing with large volumes of unstructured, user-generated text.

Theories \cite{chen2019sentiment} suggest that financial bubbles are often driven by social media hype and the intensity of contagion among users. 
As shown in Figure \ref{fig:intro}, posts from highly influential personalities often cause a growing chain reaction leading to a short squeeze and the creation of a bubble.
However, analyzing such large volumes of chaotic texts poses several challenges \cite{sawhney-etal-2021-fast}.  
Market moving events, as shown in Figure \ref{fig:intro} are rare \cite{zhao2010power} and their impact on market variables exhibit power-law distributions \cite{plerou2004origin,malevergne2005empirical}.
This power-law dynamics in online streams indicate the presence of scale free and hierarchical properties in the time domain \cite{10.1145/3397271.3401049}.

Existing works \cite{hu2018listening,stockembacl2020} that adopt RNN methods to model financial text (typically equities) do not factor in the inherent scale-free nature in online streams leading to distorted representations~\citep{NEURIPS2018_dbab2adc}.
Advances in hyperbolic learning  \cite{shimizu2021hyperbolic} motivates us to use the hyperbolic space, which better represents the scale-free nature of online text streams.
Further, online texts pose a diverse influence on cryptoprices based on their content \cite{KRAAIJEVELD2020101188,beck2019sensing}, for example, posts from
a reliable source influences future trends, as opposed to noise like vague comments as shown in Figure \ref{fig:intro}.

Building on these prospects, we present CryptoBubbles, a novel multi-bubble forecasting task (\textbf{\cref{sec:formulation}}), and a dataset comprising tweets, financial data, and speculative bubbles  (\textbf{\cref{sec:data}}) along with hyperbolic methods to model the intricate power-law dynamics associated with crypto and online user behavior pertaining to stock markets. 

Our contributions can be summarized as:
\begin{itemize}[leftmargin=12pt]
\item We formulate CryptoBubbles, a novel multi-span prediction task and publicly release a text dataset of 400+ crypto from over seven exchanges spanning over five years accompanied by over 2 million tweets.\footnote{Code and Data at: \url{https://github.com/gtfintechlab/CryptoBubbles-NAACL}}
    
\item We explore the power-law dynamics that exist in such user-generated text streams, and propose \textbf{\textsc{mbhn}}: \textbf{M}ulti \textbf{B}ubble \textbf{Hyperbolic} \textbf{N}etwork (\textbf{\cref{sec:method}}) along with other hyperbolic baselines which leverages the Riemannian manifold to model the intricate dynamics associated with crypto. 
     
\item We curate and release a test set corresponding to over 25,000 Reddit posts of 29 meme-stocks that show similar speculative user-driven dynamics (\textbf{\cref{sec:data}}) for evaluating \textsc{mbhn} under zero-shot settings (\textbf{\cref{subsec:zero_shot_dataset}}) across asset classes (Crypto $\rightarrow$ Equities) and social media platforms (Twitter $\rightarrow$ Reddit).
    
\item Through ablative (\textbf{\cref{sec:ablation}}) and qualitative experiments (\textbf{\cref{subsec:qa}}) we show the practical applicability of CryptoBubbles and hyperbolic learning.
We find that \textsc{mbhn} generalizes to cold-start scenarios (\textbf{\cref{subsec:zero_shot}}) and provides qualitative insights.
\end{itemize}

\begin{table*}[t]
\centering
\begin{minipage}[b]{0.27\textwidth}
\centering
\small
\centering
\setlength{\tabcolsep}{2.1pt}
    \begin{tabular}{lr}
        \toprule
        \textbf{Dataset Statistics}                 &         \\ \midrule
        No. of Coins   &    456    \\
        No. of Bubbles &      1,869     \\
        Len of Bubble in days   & 6.76 \text{\tiny$\pm$ 6.96} \\
        \bottomrule
    \end{tabular}
\captionof{table}{Overall CryptoBubbles data statistics}
\label{tab:overall_stat}
\end{minipage}%
\hspace{0.02\textwidth}%
\begin{minipage}[b]{0.38\textwidth}
\centering
\includegraphics[height=3.6cm]{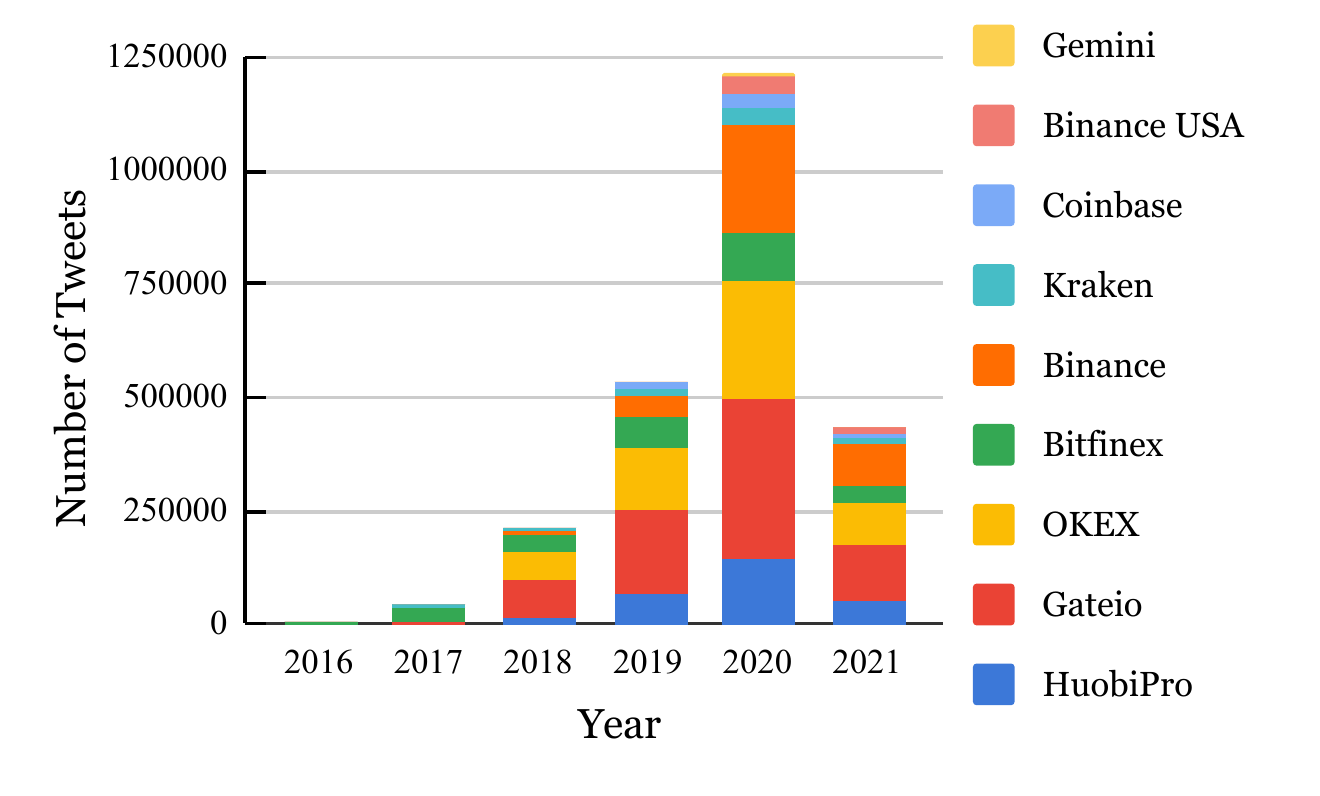}
\captionof{figure}{Yearly exchange-wise Tweet activity statistics}
\label{fig:bar_chrt}
\end{minipage}%
\hspace{0.02\textwidth}%
\begin{minipage}[b]{0.30\textwidth}
\centering
\includegraphics[height=3.7cm]{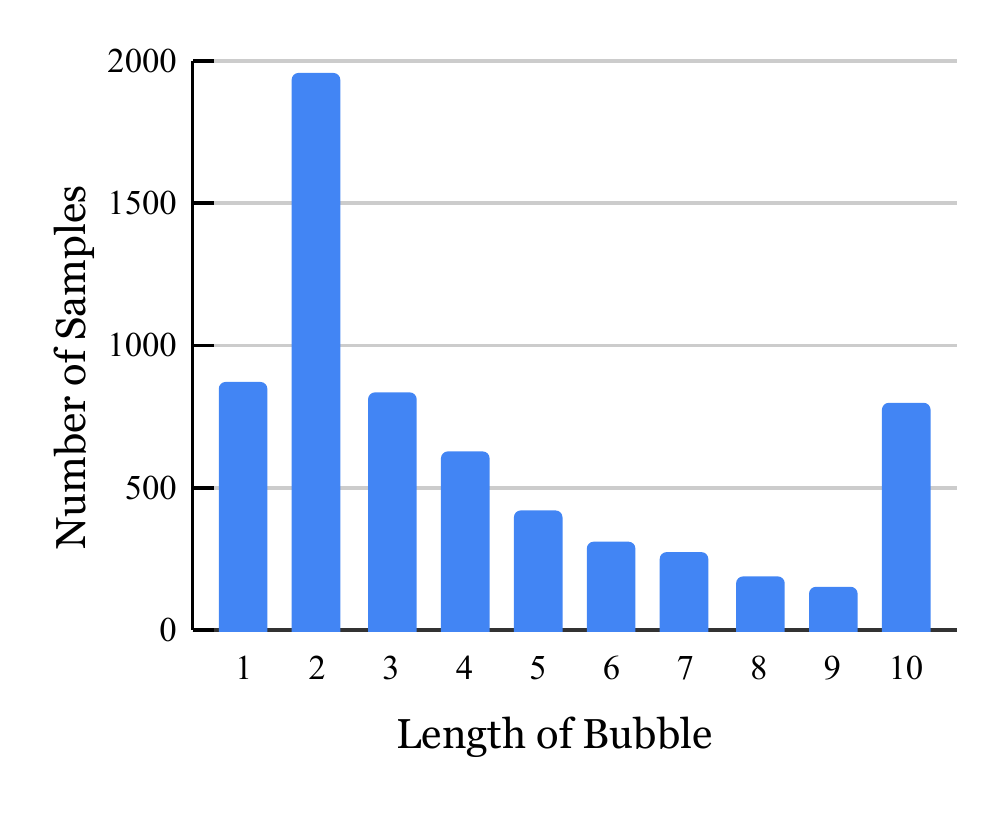}
\captionof{figure}{Frequency distribution over length of bubbles}
\label{fig:bubble_length}
\end{minipage}%
\end{table*}
\section{Background and Related Work}

\paragraph{\textbf{Cryptocurrency and Financial Bubbles}} 
Cryptocurrencies are recent digital assets that have been in use since 2008 \cite{nakamoto2008bitcoin} and rely on distributed cryptographic protocols, rather than a centralized authority to operate \cite{krafft2018experimental}. 
These assets significantly differ from traditional equities \cite{difference} which have been in use since the 17\textsuperscript{th} century \cite{sobel2000big} and have very distinct risk-return trade-offs \cite{chuen2017cryptocurrency}.
Recently, crypto trading gained popularity \cite{popular} given their low fees \cite{acess}, easy access \cite{Chepurnoy2019}, and high profit \cite{bunjaku2017cryptocurrencies} 
However, crypto trading is very challenging since it shows high volatility \cite{aloosh2020psychology}, power-law bubble dynamics \cite{fry2016negative} and effect other markets \cite{andrianto2017effect}.

\paragraph{\textbf{Financial NLP}} 
Conventional forecasting methods rely on numeric features like historical prices \cite{kohara1997stock}, and technical indicators \cite{SHYNKEVICH201771}.
These include continuous \cite{andersen2007efficient}, and neural approaches \cite{enhace_adv_lstm}.
Despite their improvements, a limitation is that they do not account for price influencing factors from text~\cite{lee-etal-2014-importance}.
Recent methods are based on the Efficient Market Hypotheses \cite{Malkiel1989} which leverage language from
earnings calls \cite{MDRMqin2019you}, online news \cite{stockembacl2020} and social media \cite{tabarietal2018causality}. 
However, a limitation is that they focus on simple assets like equities \cite{mansf} and  formulate forecasting as regression \cite{kogan-etal-2009-predicting} or ranking \cite{sawhney-etal-2021-fast} task. 
Such simple methods do not scale to highly stochastic crypto \cite{Liu2019} that show power-law dynamics \cite{KYRIAZIS2020101254}.

\paragraph{\textbf{Hyperbolic Learning}} has proved to be effective in representing  power-law dynamics \cite{ganea2018hyperbolic}, hierarchical relations \cite{Aldecoa_2015} and scale-free characteristics \cite{chami2019hyperbolic}. 
 Hyperbolic learning has been applied to various NLP \cite{dhingra-etal-2018-embedding}, and computer vision tasks \cite{khrulkov2020hyperbolic}.
However, modeling crypto texts is complex as crypto prices are driven by social media hype \cite{hyper_crypto_rw} and involve influential users \cite{ante2021elon}.
The intersection of modeling scale-free financial text streams with hyperbolic learning presents
an underexplored yet promising research avenue.

\section{Problem Formulation}\label{sec:formulation}
Let $C\!=\!\{c_1\dots,c_N\}$ denote a set of $N$ cryptos, where for each crypto $c_i$ there is an associated closing price $p_t^i$ on day $t$. 
Following \cite{Phillips2019}, we define the bubble $w$ for each crypto $c_i$ in the lookahead $T$ as the period characterized by rapid price escalation. Formally,
the logarithmic price change  $(\log p_t^i\!-\!\log p_{t-1}^i)$ takes the form,
\begin{equation}\small
    \log p_t^i - \log p_{t-1}^i =\left\{
  \begin{array}{@{}ll@{}}
    - L_t + \epsilon_t, & \text{Bubble burst}\\
     L_t + \epsilon_t, & \text{Bubble boom}\\
  \end{array}\right.
\end{equation}
where, $\epsilon_t$ are martingale difference innovations and $L_t$ given by $L_t = Lb_t$. Where, $L$ measures the shock intensity and $b_t$ is uniform over a small negative quantity $-\epsilon$ to unity.
We formulate financial bubble prediction as a multi-span prediction task since a variable number of bubbles can exist during a lookahead period $T$.
Our goal is to predict all such bubble periods $w$ in the lookahead $T$. 
Formally, given a variable number of historic texts $[d_1, \dots, d_{\tau}]$ (or other sequences like prices) for each crypto $c_i$ over a lookback of $\tau$ days, \textsc{mbhn} first outputs the number of bubbles $B$ to support multi-bubble prediction. 
Followed by identifying $B$ non-overlapping bubble periods. 

\begin{table}[t]
\centering
\small
\setlength{\tabcolsep}{2.4pt} 
\begin{tabular}{lrrr}
\toprule
 & \textbf{Train} & \textbf{Validation} & \textbf{Test} \\ \midrule
Date Range          &  03/16 - 07/20              &       07/20 -  12/20             &    12/20 - 04/21           \\
Split Ratio    & 48\%        & 29\%             & 23\%       \\
No. of Coins   & 307            & 357                 & 364           \\
No. of Tweets  & 1.15 M        & 730 K              & 561 K        \\
\% Bubbles     & 8.27\%         & 7.93\%              & 23.30\%       \\ \bottomrule
\end{tabular}
\caption{Sample level CryptoBubble dataset statistics along with its chronological date splits.}
\label{tab:sample_stats}
\end{table}

\section{Dataset Curation and Processing\footnote{We provide extensive data creation steps along with details on bubble creation in the Appendix.}}\label{sec:data}

\paragraph{\textbf{Data Mining}} 
To create CryptoBubbles dataset, we select the top 9 crypto exchanges such as Binance, Gateio, etc and choose around 50 most traded cryptos by volume from each exchange and obtain 456 cryptos in total.
For these cryptos we mine 5 years of daily price data consisting of opening, closing, highest and lowest prices from 1\textsuperscript{st} Mar'16 to 7\textsuperscript{th} Apr'21 using CryptoCompare.\footnote{\url{https://www.cryptocompare.com/}} 
Next, we extract crypto related tweets under Twitter's official license.  
Following \cite{stocknet}, we extract crypto-specific tweets by querying regex ticker symbols, for instance, “$\text{\$DOGE}\backslash$\text{b}” for \textit{DogeCoin}. 
We mine tweets for the same date range as the price data and obtain  2.4 million tweets. 
We detail the yearly number of tweets from each exchange in Figure \ref{fig:bar_chrt} and observe that the number of tweets increases every year, indicating the growing popularity of speculative trading via social media. 

\paragraph{\textbf{Bubble Creation}} 
To identify bubbles, we use the PSY model \cite{phillips2015testing} which is a widely used bubble detection method in financial time-series analysis \cite{cheung2015crypto,harvey2016tests}.
Following \cite{corbet2018datestamping} we feed the closing prices of each asset $c_i$ to the PSY model, which outputs date spans for each bubble.
To further review the ground-truth annotations produced by the PSY model, all annotations were reviewed by five experienced financial analysts achieving a Cohen's $\kappa$ of 0.93. We find that the reviewers agree with the annotations for 90\% of the bubbles. During reviewer disagreement (5\% bubbles), we took the majority of all annotators. For 5\% of the bubbles all reviewers agreed that the annotations were incorrect, during which we considered the annotations proposed by the analysts.  
We provide overall dataset and bubble statistics in Table \ref{tab:overall_stat} and Figure \ref{fig:bubble_length}, respectively. 

\paragraph{\textbf{Sample Generation}}
To generate data samples, we use a sliding window of length $\tau$ and consider all texts posted within this window for making predictions over the next $T$ days. 
Next, we temporally split all the data samples into train, validation, and test as shown in Table \ref{tab:sample_stats}.
We note that our data is heavily skewed; our evaluation set contains new cryptos and spans the COVID-19 period suggesting that CryptoBubbles is challenging. 

\begin{figure*}[t]
\centering \includegraphics[width=\textwidth]{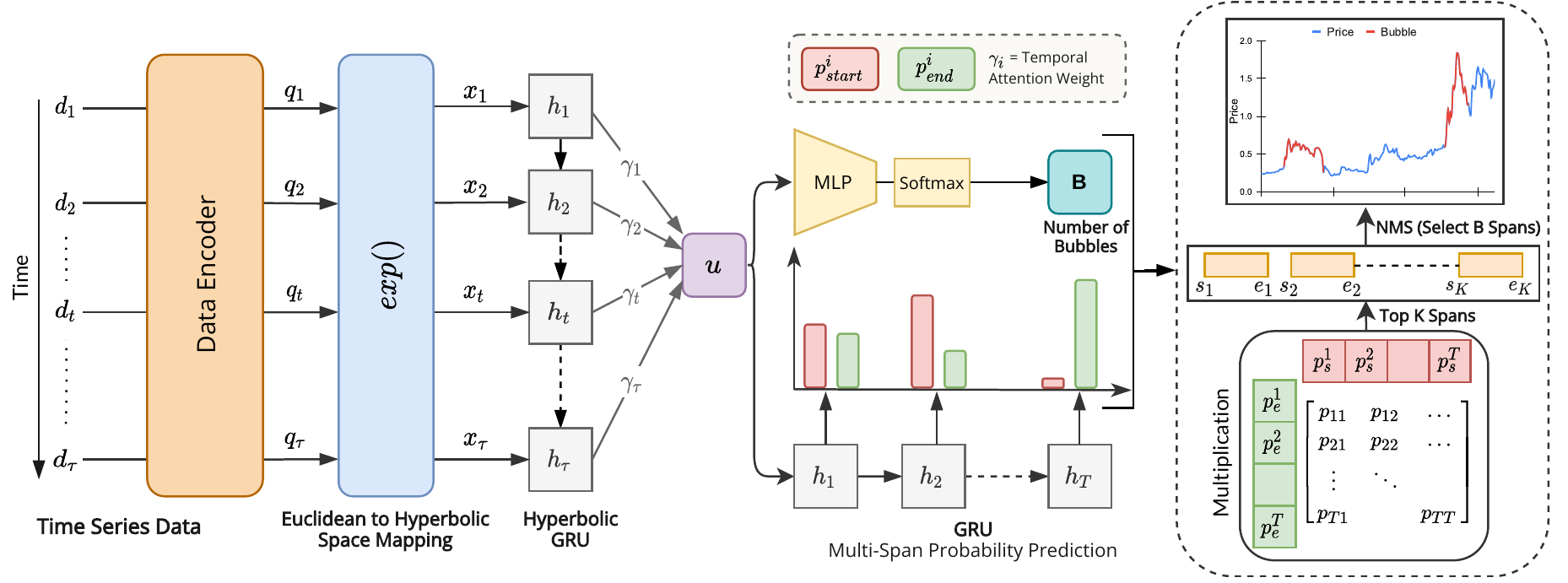}
\caption{An overview of \textsc{mbhn}, hyperbolic mappings, hyperbolic GRU, and multi-span extraction. Data Encoder is used to encode any temporal sequence (historic prices or texts). The data encoder in case of texts is BERT.}
\label{fig:overview}
\end{figure*}

\subsection{Zero Shot Reddit Data} \label{subsec:zero_shot_dataset}
We curate zero-shot Reddit data to test our model's ability to generalize across different asset classes and social media platforms. We analyze 12 meme cryptos and 17 meme equities (for instance, GameStop, and DOGE) selected based on social media activity over 15 months from 15\textsuperscript{th} Jan'20 to 3\textsuperscript{rd} April'21. 
We mine Reddit posts and comments from top trading subreddits such as r/wallstreetbets using the PushshiftAPI.\footnote{\url{https://github.com/pushshift/api}}
We scrape daily price data using Yahoo Finance for equities and CoinGecko for crypto and use the PSY model \cite{phillips2015testing} to identify bubbles. 
We follow the same process used for Twitter and summarize the statistics in Table \ref{tab:reddit_st}.
We note that zero-shot data establishes a challenging environment for evaluating CryptoBubbles since it contains varied post lengths and unseen assets. 
\begin{table}[t]
\small
\setlength{\tabcolsep}{3.5pt}
\begin{tabular}{@{}lrr@{}}
\toprule
  & \textbf{Cryptocurrency} & \textbf{Equity} \\ \midrule
No. of Posts               & 30.69 ${\scriptstyle\pm}$ 29.48    & 30.08 ${\scriptstyle\pm}$ 30.26   \\
No. of Tokens per post     & 34.50 ${\scriptstyle\pm}$ 34.87  & 23.06 ${\scriptstyle\pm}$ 23.62 \\
No. of Bubbles             & 0.16 ${\scriptstyle\pm}$ 0.42    & 0.22 ${\scriptstyle\pm}$ 0.45    \\
Length of Bubbles in days         & 4.28 ${\scriptstyle\pm}$ 2.73    & 4.60 ${\scriptstyle\pm}$ 2.93      \\ \bottomrule
\end{tabular}
\caption{Zero-shot Reddit data statistics for cryptocoins and equities along with their bubble statistics.}
\label{tab:reddit_st}
\end{table}

\section{Methodology: \textsc{mbhn}}\label{sec:method}
\subsection{Preliminaries on the Hyperbolic Space}\label{subsec:hyp_space}
We implement \textsc{mbhn} on the Poincaré ball model, defined as $(\mathcal{B},g_x^{\mathcal{B}})$,  where the manifold $\mathcal{B}\!=\! \{\boldsymbol{x}\in\mathbb{R}^n: ||\boldsymbol{x}||<1\}$, with the Riemannian metric $g_{\boldsymbol{x}}^\mathcal{B}\! =\! \lambda_{\boldsymbol{x}}^2g^E$, where the conformal factor $\lambda_{\boldsymbol{x}} \!=\! \frac{2}{1-||\boldsymbol{x}||^2}$ and $g^E\! =\! \text{diag}[1, .. , 1]$ is the Euclidean metric tensor. 
We denote the tangent space centered at point $\boldsymbol{x}$ as $\mathcal{T}_{\boldsymbol{x}}\mathcal{B}$.
Following \cite{Ungar01hyperbolictrigonometry}, we generalize Euclidean operations to the hyperbolic space using the Möbius operations. 

\paragraph{\textbf{Möbius Addition}} $\oplus$ for a pair of points $\boldsymbol{x},\boldsymbol{y}\!\in\!\mathcal{B}$, 
\begin{equation}
    \boldsymbol{x}\oplus \boldsymbol{y} \!=\! \frac{(\!1\!+\! 2\langle \boldsymbol{x\!,\!y} \rangle \!+\! ||\boldsymbol{y}||^2)\boldsymbol{x}\!+\!(1\!-\!||\boldsymbol{x}||^2)\boldsymbol{y}}{1\! +\! 2\langle\boldsymbol{x,y}\rangle \!+\! ||\boldsymbol{x}||^2||\boldsymbol{y}||^2}
\end{equation}
where, $\langle.,.\rangle$ denotes the inner product and $||\cdot||$ denotes the norm.
To perform operations in the hyperbolic space, we define the exponential and logarithmic map to project Euclidean vectors to the hyperbolic space, and vice versa. 

\paragraph{\textbf{Exponential Mapping}} maps a tangent vector $\boldsymbol{v}\in\mathcal{T}_{\boldsymbol{x}}\mathcal{B}$ to a  point $\text{exp}_{\boldsymbol{x}}(\boldsymbol{v})$ in the hyperbolic space,
\begin{equation}
    \text{exp}_{\boldsymbol{x}}(\boldsymbol{v}) = \boldsymbol{x}\oplus\left(\text{tanh}\left(\frac{||\boldsymbol{v}||}{1-||\boldsymbol{x}||^2}\right)\frac{\boldsymbol{v}}{||\boldsymbol{v}||}\right)
\end{equation}

\paragraph{\textbf{Logarithmic Mapping}} maps a point $\boldsymbol{y}\in\mathcal{B}$ to a point $\text{log}_{\boldsymbol{x}}(\boldsymbol{y})$ on the tangent space at $\boldsymbol{x}$,
\begin{equation}\small
    \text{log}_{\boldsymbol{x}}({\boldsymbol{y}}) \!=\! (1-||\boldsymbol{x}||^2)\text{tanh}^{-1}\left(\!||\!-\!\boldsymbol{x}\oplus\boldsymbol{y}||\!\right)\!\frac{-\boldsymbol{x}\oplus\boldsymbol{y}}{||\!-\!\boldsymbol{x}\!\oplus\!\boldsymbol{y}||}
\end{equation}

\paragraph{\textbf{Möbius Multiplication}} $\otimes$ multiplies features $\boldsymbol{x}\in\mathcal{B}^{C}$ with matrix $\boldsymbol{W}\in\mathbb{R}^{C'\times C}$, defined as,
\begin{equation}
    \boldsymbol{W}\otimes\boldsymbol{x} =\text{exp}_{\boldsymbol{o}}(\boldsymbol{W} \text{log}_{\boldsymbol{o}}(\boldsymbol{x}))
\end{equation}
We present an overview of \textsc{mbhn} in Figure \ref{fig:overview}.
We first explain temporal feature (price or text) extraction using hyperbolic GRU and the temporal attention (\textbf{\cref{subsec:ta}}), followed by the decoder architecture to predict multiple bubble spans (\textbf{\cref{subsec:spans}}).

\subsection{Hyperbolic Temporal Encoder}\label{subsec:ta}
Temporal dependencies in crypto data show power-law dynamics and scale-free nature \citep{zhang2018some,gabaix2003theory}.
As shown in Figure \ref{fig:overview} to capture the hierarchical and temporal dependencies in temporal crypto data (online texts or asset prices), we implement a Gated Recurrent Unit (GRU) in the hyperbolic space \citep{NEURIPS2018_dbab2adc} which better models the scale-free dynamics of online streams \citep{NIPS2019_8733}.

As shown in Figure \ref{fig:overview}, the input to \textsc{mbhn} can be any temporal sequence $[d_1,\dots,d_{\tau}]$ such as prices or texts. We use a data encoder to encode each item $d_t$ as features $q_t$. Specifically, we use Bidirectional Encoder Representations from Transformers (BERT) \cite{devlin-etal-2019-bert} for encoding texts as $q_t=\text{BERT}(d_t)$. While we feed raw price vectors comprising of a cryptos’s 
closing price $p_c^t$, highest price $p_h^t$ and lowest price $p_l^t$ for day $t$ for encoding historic prices as $q_t=[p_c^t,p_h^t,p_l^t]$. For encoding both prices and texts together we concatenate the price and text features.  
To apply the hyperbolic operations, we first map these Euclidean features $\boldsymbol{q}_t$ to hyperbolic features $\boldsymbol{x}_t\in\mathcal{B}^C$ for each crypto $c_i$ via the exponential map, given by,
$
    \boldsymbol{x}_t \!=\! \text{exp}_{\boldsymbol{o}}(\boldsymbol{q}_t)
$.
We define the hyperbolic GRU on features $\boldsymbol{x}_t$ as,

\footnotesize
\begin{align*}
    \boldsymbol{z}_t\!&=\! \sigma \text{log}_{\boldsymbol{o}}\!\left(\!\boldsymbol{W}^z\! \otimes\! \boldsymbol{h}_{t-1} \!\oplus\! \boldsymbol{U}^z\! \otimes\! \boldsymbol{x}_t \!\oplus \!\boldsymbol{b}^z\!\right)\tag{\footnotesize{Update gate}}\\
    \boldsymbol{r}_t \!&= \!\sigma\text{log}_{\boldsymbol{o}}\!\left(\!\boldsymbol{W}^r \!\otimes \!\boldsymbol{h}_{t-1}\!\oplus\! \boldsymbol{U}^r\!\otimes\! \boldsymbol{x}_t \!\oplus\! \boldsymbol{b}^r\!\right) \tag{\footnotesize{Reset gate}}\\
    \Bar{\boldsymbol{h}}_t \!&= \!\psi^\otimes\!\left(\! \!\boldsymbol{W}^h\!\text{diag}\!(\!\boldsymbol{r}_t\!)\! \otimes\! \boldsymbol{h}_{t-1} \!\oplus \!\boldsymbol{U}^h\!\otimes\! \boldsymbol{x}_t \!\oplus\! \boldsymbol{b}^h\!\right) \text{\footnotesize{(Current state)}}\\
    \boldsymbol{h}_t &= \boldsymbol{h}_{t-1} \oplus \text{diag}(\boldsymbol{z}_t) \otimes (-\boldsymbol{h}_{t-1} \oplus \Bar{\boldsymbol{h}_t}) \tag{\footnotesize{     Final state}}
\end{align*}
\normalsize
where, $\Psi^\otimes$ denotes hyerbolic non-linearity \citep{NEURIPS2018_dbab2adc} and $\boldsymbol{W,U,b}$ are learnable weights. 
We denote the Hyperbolic GRU as $\text{HGRU}(\cdot)$ which takes temporal crypto features $\boldsymbol{X}= \{\boldsymbol{q}_1,\dots,\boldsymbol{q}_{\tau c_i}\}$ as input and  outputs features $\boldsymbol{K} = \{h_1,\dots,h_{\tau c_i}\}$, which is the concatenation of all hidden states $\boldsymbol{h}_t$ of crypto $c_i$, given by,
\begin{equation}
    \boldsymbol{K} = \text{HGRU}(\exp_{\boldsymbol{o}}(\boldsymbol{X}))
\end{equation}
All texts (or prices) released in the lookback $\tau$ may not be equally informative and have \textit{diverse influence} over a crypto's trend \cite{KRAAIJEVELD2020101188}.
We use a temporal attention mechanism \cite{luong-etal-2015-effective} to emphasize streams likely to have a substantial influence on the bubble. 
The attention mechanism learns attention weights $\{\gamma_j\}_{j=1}^{\tau c_i}$ to adaptively aggregate hidden states of $\text{HGRU}$ into a \textit{crypto information vector} $\boldsymbol{u}_i$,
\begin{gather}
\boldsymbol{u}_{i} = \sum_j \gamma_{j}\log_{\boldsymbol{o}}({\boldsymbol{ h}}_{j})\\
\gamma_{j} = \text{Softmax}_j(\log_{\boldsymbol{o}}({h}_{j})^\text{T}(\boldsymbol{W}\log_{\boldsymbol{o}}(\boldsymbol{K})))
\end{gather}
where, $\boldsymbol{W}$ are learnable weights.

\definecolor{forestgreen}{rgb}{0.13, 0.55, 0.13}
\definecolor{royalblue}{rgb}{0.25, 0.41, 0.88}
\definecolor{blueviolet}{rgb}{0.54,0.17,0.89}
\begin{table*}[t]
\centering
\begin{minipage}[b]{0.26\textwidth}
\centering
\includegraphics[height=3cm]{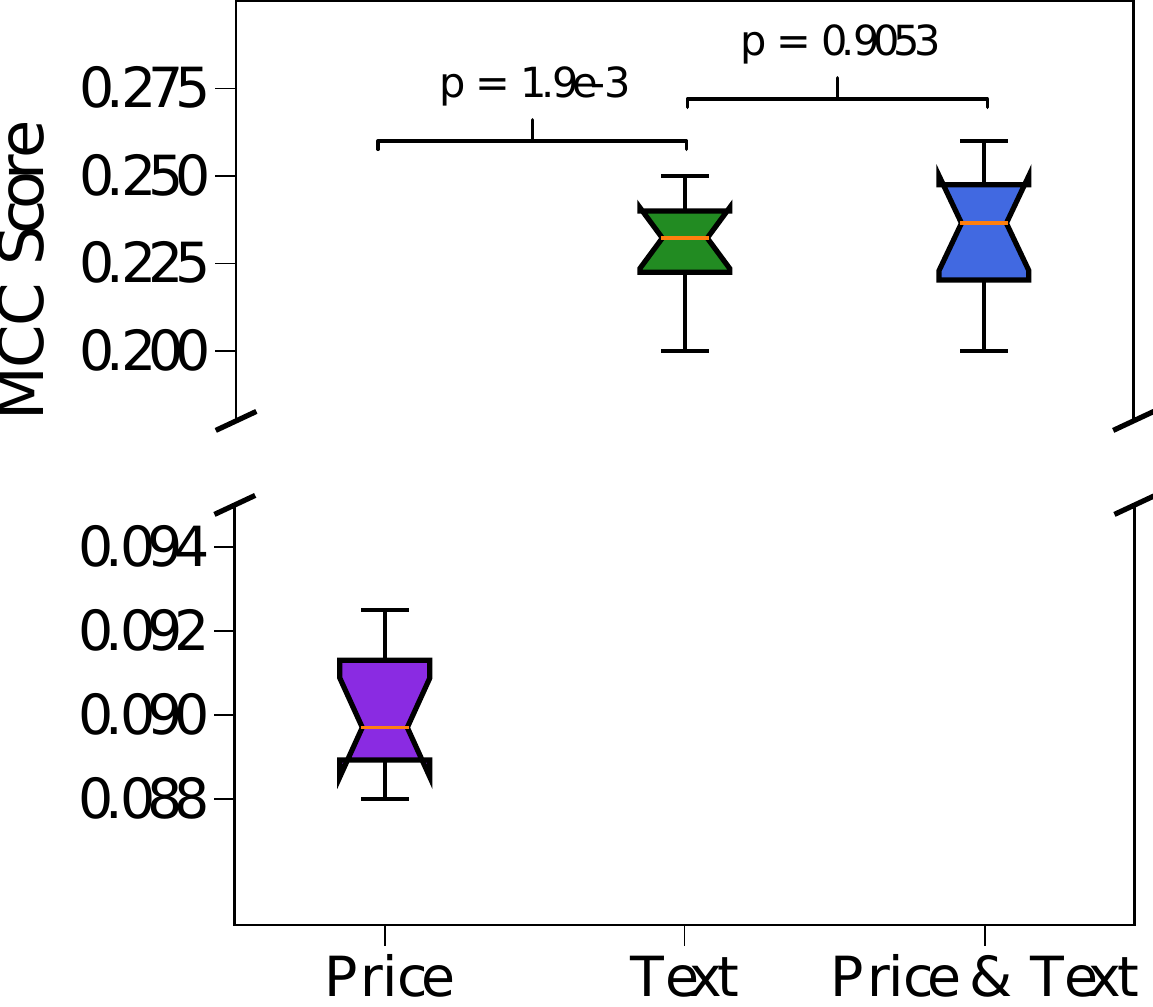}
\captionof{figure}{Distribution of \textbf{\textsc{mbhn}} performance on bubble span prediction with price (\textcolor{blueviolet}{violet}), text (\textcolor{forestgreen}{green}) and price+text (\textcolor{royalblue}{blue}) inputs along with confidence intervals.}
\label{fig:boxplot}
\end{minipage}%
\hspace{0.02\textwidth}%
\begin{minipage}[b]{0.72\textwidth}
\centering
\small
\centering
\setlength{\tabcolsep}{3.5pt}

    \begin{tabular}{lrrrrr}
        \toprule
        \textbf{}            & \multicolumn{5}{c}{\textbf{Task}}                                                                                                                                         \\ \cmidrule(l){2-6} 
        \textbf{Model}       & \multicolumn{3}{c}{\textbf{Bubble Span}}                                                             & \multicolumn{2}{c}{\textbf{Number of Bubbles}}                     \\ \cmidrule(l){2-6} 
        \textbf{}            & \multicolumn{1}{c}{\textbf{F1$\uparrow$}} & \multicolumn{1}{c}{\textbf{MCC$\uparrow$}} & \multicolumn{1}{c}{\textbf{EM$\uparrow$}} & \multicolumn{1}{c}{\textbf{Acc (\%)$\uparrow$}} & \multicolumn{1}{c}{\textbf{F1$\uparrow$}} \\ \midrule
        EGRU & $0.48 {\scriptstyle\pm 2e-4 }$ & $0.15 {\scriptstyle\pm 6e-5}$   & $0.47 {\scriptstyle\pm 3e-4}$   & $56.70 {\scriptstyle\pm 2e-4} $ & $0.22 {\scriptstyle\pm 1e-2}$                                 \\
        EGRU+Attn         & $0.50 {\scriptstyle\pm 3e-5  }$& 0.18 ${\scriptstyle\pm 1e-4}$ & $0.49 {\scriptstyle\pm 5e-4   }$ & $59.48 {\scriptstyle\pm 6e-5  }$ & 0.25 ${\scriptstyle\pm 3e-4}$                                \\
         HGRU       & \textit{0.52$\textsuperscript{*\dag}$ ${\scriptstyle\pm 1e-4}$} & \textit{0.20\textsuperscript{*\dag} ${\scriptstyle\pm 3e-4}$} & \textit {0.50\textsuperscript{*\dag} ${\scriptstyle\pm 4e-4}$ }  & \textit{60.32\textsuperscript{*\dag} ${\scriptstyle\pm 2e-4}$ }  & \textit{0.27\textsuperscript{*\dag} ${\scriptstyle\pm 3e-4}$} \\
        
        \textbf{MBHN} & \textbf {0.53 $\textsuperscript{*\dag} {\scriptstyle\pm 1e-3} $}  & \textbf{0.23 $\textsuperscript{*\dag} {\scriptstyle\pm 4e-4}$} & \textbf{0.52 $\textsuperscript{*\dag} {\scriptstyle\pm 2e-3}$}   & \textbf{62.01 $\textsuperscript{*\dag} {\scriptstyle\pm 3e-4}$ }  & \textbf{0.29\textsuperscript{$*\dag$} $\scriptstyle\pm 1e-3$}           \\                    
        \bottomrule
    \end{tabular}
\captionof{table}{Ablation over \textbf{\textsc{mbhn}} components (over 10 independent runs). \textbf{Bold}, \textit{italics} represent \textbf{best} and \textit{second-best} results, respectively. $*$ and $\dag$ indicate significant ($p<0.01$) improvement over EGRU (Euclidean GRU) and EGRU+Attn (Euclidean GRU with Attention), respectively under Wilcoxon's Signed Rank Test.}
\label{tab:ablation}
\end{minipage}%
\end{table*}

\subsection{Multi-Span Extraction and Optimization}\label{subsec:spans} 
\paragraph{\textbf{Span Prediction}}
To extract the bubble spans in the lookahead $T$, we take inspiration from \cite{sutskever2014sequence} and use a GRU based decoding network.
We use the outputs of the encoder $\boldsymbol{u}_i$ for each crypto $c_i$  as the initial state of the GRU.
For each day $t \in [1,\dots,T]$, we predict the probability of day $t$ being a starting or an ending day denoted by $p_{\text{start}}^{t}$ and $p_{\text{end}}^{t}$ respectively, given by,
\begin{align}
    p_{\text{start}}^{t} &= \text{Sigmoid}(\boldsymbol{W}^S\text{GRU}(\boldsymbol{u}_i))\\
    p_{\text{end}}^{t} &= \text{Sigmoid}(\boldsymbol{W}^E\text{GRU}(\boldsymbol{u}_i))
\end{align}
where $\boldsymbol{W}^S$,$\boldsymbol{W}^E$ are learnable weights. 

\paragraph{\textbf{Multi-Span Extraction}}
To identify multiple bubble spans, we first predict the number of bubbles $B$ for each cryptocoin $c_i$ in the lookahead $T$ and model it as a classification task, given by,  
\begin{equation}
    B = \text{argmax}(\text{Softmax}(\text{MLP}(\boldsymbol{u}_i)))
\end{equation}
As shown in Figure \ref{fig:overview}, we extract $B$ non-overlapping bubbles for each coin $c_i$ using the non-maximum suppression
(NMS) algorithm \cite{rosenfeld1971edge}. 
For the $j^{th}$ bubble our model predicts a starting index $s_j$ and an ending index $e_j$, where $s_j\! <\! e_j \!\leq\! T$, $j \in [1, \dots, B]$.
We first extract \textit{top-K} bubble spans $\mathbf{S}$ based on the bubble scores, calculated as $p_{start}^{s_j} \times p_{end}^{e_j}$ for the $j^{\text{th}}$ bubble  starting at index $s_j$ and ending at index $e_j$.
Next, we initialize an empty set $\mathbf{\overline{S}}$ and add the bubble $(s_j, e_j)$ that posses the maximum bubble score to the set $\mathbf{\overline{S}}$, and remove it from the set $\mathbf{S}$.
We also delete any remaining bubble spans $(s_k, e_k)$ having an overlap with bubble $(s_j, e_j)$. 
This process is repeated for all the remaining spans in $\mathbf{S}$, until $\mathbf{S}$ is empty or we get $B$ bubble spans in the set $\mathbf{\overline{S}}$.

\paragraph{\textbf{Network Optimization}}
We optimize \textsc{mbhn} using a combination of losses corresponding to the three tasks; 1) Bubble start date prediction, 2) Bubble end date prediction and 3) Number of bubble prediction. 
We optimize \textsc{mbhn} using binary cross entropy loss over both bubble start date and end date prediction.
For optimizing over the number of bubbles, we use the Focal loss \cite{lin2017focal}. 
The net loss is the sum of the three losses.

\section{Experimental Setup}
\paragraph{\textbf{Preprocessing}}
We pre-process English tweets
using NLTK (Twitter mode), for treatment of URLs,
identifiers (@) and hashtags (\#). We adopt the BertTokenizer for tokenization and use the pre-trained BERT-base-cased model for text embeddings. 
We align days by dropping samples that lack tweets for a consecutive 5-day lookback
window. 
\paragraph{\textbf{Training Setup}}
We adopt grid search to find optimal hyperparameters based on the validation MCC for all methods. 
For NMS, we use an overlapping threshold of 2 units.
We use default $\alpha= 2,\gamma= 5$ for the focal loss and learning rate $\in(1e^{-5},1e^{-3})$ to train the models using the Adam optimizer.
\paragraph{\textbf{Evaluation Metrics}}
We evaluate all methods using \textbf{F1}-score, Mathews Correlation Coefficient (\textbf{MCC}) and Exact Match (\textbf{EM}) (We provide metric equations in \textit{Appendix}).
On the bubble span level, we use F1 and MCC, which measure the overlap between the predicted and the true bubble spans. 
While EM  measures the percentage of
overall predicted bubbles that exactly match the
true bubble spans.
We also evaluate on the ``number of bubble'' task via \textbf{F1}-Score and Accuracy (\textbf{Acc}).

\subsection{Baseline Methods}
We use BERT for text encoding in all models. Each baseline forecasts prices over lookahead $T$ and uses the PSY model to detect bubbles.
\begin{itemize}[leftmargin=12pt]
    \item \textbf{ARIMA }Auto-regressive moving average based model that uses past prices from 100 days as input for forecasting \cite{yenidougan2018bitcoin}.

\item \textbf{W-LSTM }A LSTM model with autoencoders that encode noise-free data obtained via wavelet transform of historic prices \cite{bao2017deep}.
\item\textbf{A-LSTM } Adversarially trained LSTM on price inputs for forecasting \cite{ijcai2019-0810}.
\item \textbf{Chaotic }A Hierarchical GRU model applied on texts within and accross days \cite{hu2018listening}.
\item \textbf{MAN-SF(T) }Hierarchical attention on texts within and across days \cite{sawhney2020deep}.
\item \textbf{CH-RNN } An RNN coupled with cross-modal attention on price and texts \cite{wu2018hybrid}.
\item \textbf{SN - HFA: } StockNet - HedgeFundAnalyst, a variational autoencoder with attention on texts and prices \cite{xu2018stock}.

\end{itemize}

\section{Results and Analysis}
\subsection{Ablation Study}\label{sec:ablation}
Through Figure \ref{fig:boxplot} we observe that augmenting price-based \textsc{mbhn} with financial texts leads to significant ($p<0.01$) improvements, suggesting the importance of leveraging textual sources to forecast the formation of speculative bubbles.
This observation ties up with \cite{sawhney-etal-2021-fast} who show that online texts often indicate market surprises that may not be well captured by prices alone.
Noting insignificant ($p>0.01$) improvements on using both price and text information, we further probe \textsc{mbhn} performance benefits from each of its components in Table \ref{tab:ablation} with only textual inputs.
We note the biggest improvements on replacing the Euclidean encoders with their hyperbolic variants, suggesting that the inherent power-law dynamics and scale-free nature of online texts and financial bubbles are better represented in the hyperbolic space \cite{ganea2018hyperbolic}. 
Next, on adding temporal attention, we note improvements as \textsc{mbhn} can better distinguish noise inducing text from relevant market signals, minimizing false evaluations and overreactions \cite{de1990noise}.
The temporal attention mechanism can likely diminish the impact of such noise (rumors, vague comments) and better capture the diverse influence of texts.

\subsection{Performance Comparison}\label{subsec:performance}
\begin{table}[t]
\centering
\setlength{\tabcolsep}{1.5pt}

\small
\begin{tabular}{lrrr}
\toprule
\textbf{Model}       & \textbf{F1} $\uparrow$   & \textbf{MCC} $\uparrow$  & \textbf{EM} $\uparrow$  \\ \midrule
ARIMA \cite{yenidougan2018bitcoin}               & 0.02          & 0.00          & 0.02          \\
W-LSTM \cite{bao2017deep}            & {0.09}          & {0.04}          & {0.03}         \\
A-LSTM \cite{ijcai2019-0810}            & {0.45}          & {0.13}          & {0.39}         \\
Chaotic \cite{hu2018listening}            & {0.49}          & {0.14}          & {0.45}         \\
MAN-SF(T) \cite{sawhney2020deep}            & {0.49}          & {0.16}          & {0.46}         \\
CH-RNN \cite{wu2018hybrid}            & {0.50}          & {0.19}          & {0.47}         \\
SN - HFA \cite{xu2018stock}            & \textit{0.51}          & \textit{0.21}          & \textit{0.49}         \\
\textbf{\textbf{MBHN} (Ours)} & \textbf{0.53\textsuperscript{*}} & \textbf{0.25\textsuperscript{*}} & \textbf{0.53\textsuperscript{*}} \\ \bottomrule
\end{tabular}
\caption{Performance comparison against baselines (mean of 10 independent runs).  \textbf{Bold} and \textit{italics} represent \textbf{best} and \textit{second-best} results, respectively. $*$ indicates statistically significant ($p<0.01$) improvement over SN-HFA under Wilcoxon's Signed Rank Test.}
\label{tab:perf}
\end{table}
We compare the performance of \textsc{mbhn} with baselines in Table \ref{tab:perf}. 
We observe that \textsc{mbhn} through hyperbolic learning significantly outperforms ($p<0.01$) all baselines. 
This observation validates our premise of formulating CryptoBubbles as an extractive multi-span task and suggests its practical applicability in predicting speculative financial risks.
Further, we note that methods that use crypto affecting information from online texts (\textsc{mbhn}, SN-HFA) generate better performance than approaches that only use price data (A-LSTM, W-LSTM).
These improvements re-validate the effectiveness of using online texts to encode investor sentiment and market information.
Next, we note that \textsc{mbhn} requires less historical data ($\tau=5$) compared to conventional methods (ARIMA, $\tau=100$), while achieving better performance, suggesting \textsc{mbhn} applicability in low data scenarios.

\begin{table}[t]
\small
\centering
\setlength{\tabcolsep}{1.8pt}
\begin{tabular}{clrrr}
\toprule
\textbf{Asset Type}                          & \textbf{Asset Name}                & \multicolumn{1}{c}{\textbf{F1$\uparrow$}} & \multicolumn{1}{c}{\textbf{MCC$\uparrow$}} & \multicolumn{1}{c}{\textbf{EM$\uparrow$}} \\ \midrule
\multicolumn{1}{c}{\multirow{3}{*}{Equity}}                        & \multicolumn{1}{l}{Top 1: CCIV}          & 0.73                            & 0.48                             & 0.71                            \\
\multicolumn{1}{c}{}                        & \multicolumn{1}{l}{Bottom 1: PLTR}          & 0.26                            & -0.05                            & 0.17                            \\
\multicolumn{1}{c}{}                        & \multicolumn{1}{l}{All Equities} & 0.52                            & 0.07                             & 0.63                            \\ \midrule
\multicolumn{1}{c}{\multirow{3}{*}{Cryptocoin}} & \multicolumn{1}{l}{Top 1: DOGE}          & 0.54                            & 0.26                             & 0.55                            \\
\multicolumn{1}{c}{}                        & \multicolumn{1}{l}{Bottom 1: BASED}         & 0.33                            & -0.01                            & 0.43                            \\
\multicolumn{1}{c}{}                        & \multicolumn{1}{l}{All Cryptocurrencies} & 0.50                            & 0.05                             & 0.70      \\ \bottomrule                     
\end{tabular}
\caption{MBHN's performance on Reddit meme stock data under zero shot settings. Top 1 and Bottom 1 denote MBHN's best and the worst performancing assets.}
\label{tab:zero_shot}
\end{table}

\begin{figure*}[t]
\centering \includegraphics[width=\textwidth]{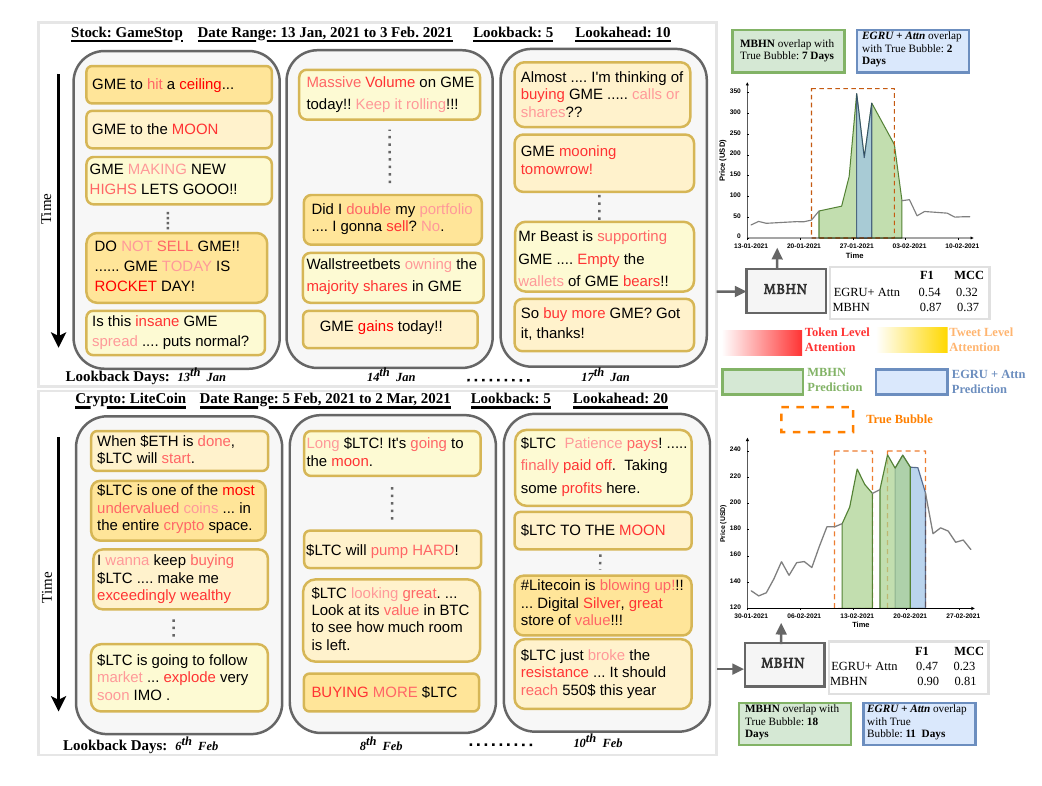}
\caption{\textbf{Top:} Reddit posts of GameStop from 13\textsuperscript{th} Jan'21 to 3\textsuperscript{rd} Feb'21. \textbf{Bottom:} Twitter tweets of LiteCoin from 5\textsuperscript{th} Feb'21 to 2\textsuperscript{nd} March'21 along with token level attention and temporal tweet level attention visualisation. We also show \textsc{mbhn}'s performance with its Euclidean variant on CryptoBubbles bubble detection task.}
\label{fig:curr}
\end{figure*}
\begin{figure}[t]
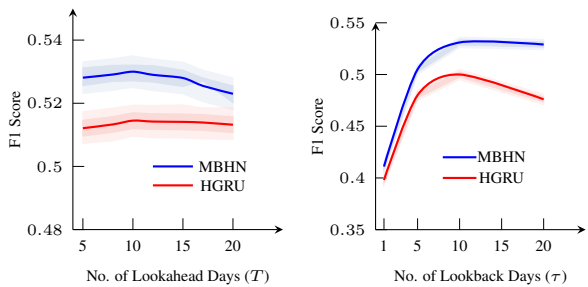

    \hspace{-0.1cm}
    \begin{subfigure}[b]{0.485\linewidth}
    \includegraphics[width=1\textwidth, height =3.8 cm]{lookahead.tikz}
    \end{subfigure}
    ~
    \begin{subfigure}[b]{0.485\linewidth}
    \includegraphics[width=1\textwidth, height =3.8 cm]{lookback.tikz}
    \end{subfigure}   
\caption{Sensitivity to parameters $T$ and ${\tau}$}
\label{fig:lookback}
\end{figure}
\subsection{Zero-shot Learning}\label{subsec:zero_shot}
To evaluate \textsc{mbhn} applicability to unseen asset classes and social media linguistic styles we test it on unseen Reddit meme-stock data (\textbf{\cref{sec:data}}) under zero-shot settings and summarise the results in Table \ref{tab:zero_shot}.
We observe that \textsc{mbhn} can generalize over the true market sentiment to an extent and transfer knowledge across the two social media by being invariant to post styles, lengths, and social media structure. 
Further, these observations suggest that \textsc{mbhn} can be leveraged for cold start scenarios (completely new assets), which is quite frequent for cryptocurrencies \cite{new_cryto}. 
We speculate \textsc{mbhn}'s transferability to hyperbolic learning due to its ability to accurately reflect complex scale-free relations between social media texts. 
This observation ties up with \cite{Khrulkov_2020_CVPR} who show that hyperbolic learning is useful under zero-shot settings. 
These observations collectively demonstrate \textsc{mbhn}'s ability to generalize across social media and asset classes, including meme stocks such as GameStop, which purely see bubbles due to social media hype \cite{chohan2021counter}.

\subsection{Sensitivity to Lookback and Lookahead}
We study the impact of historical context on \textsc{mbhn}'s performance in Figure \ref{fig:lookback} by varying the availability of historic information.
We observe lower performance on using shorter lookback periods, indicating inadequacy to capture the market influencing signals, likely as public information requires time to absorb into price fluctuations \cite{doi:10.1080/14697688.2012.672762}. 
As we increase the lookback, we find that larger periods allow the inclusion of stale information from older days having a relatively lower influence on prices \cite{bernhardt2004informed} that may deteriorate the model performance.
However, our temporal attention mechanism, to an extent, learns to filter our salient texts, which induce bubbles in the market. 
Further, we analyze \textsc{mbhn}'s sensitivity to the number of lookahead days $T$ in Figure \ref{fig:lookback}. 
We observe that our model is robust to different lookahead periods, suggesting \textsc{mbhn}'s generalizability for predicting speculative bubbles over different risk taking appetites.
\subsection{Qualitative Analysis}\label{subsec:qa}
We conduct an extended study to elucidate \textsc{mbhn}'s explainable predictions and practical applicability. As shown in Figure \ref{fig:curr}, we analyze the recent posts corresponding to LiteCoin tweets and Gamestop zero-shot Reddit posts.
 
\paragraph{\textbf{Practical applicability of CryptoBubbles and \textsc{mbhn}}}
We first calculate temporal attention scores across tweets and observe that \textsc{mbhn} is able to attend to more influential tweets to an extent for both meme cryptocurrencies and stocks under zero-shot settings providing valuable pieces of information to better judge speculative risks \cite{kumar2009gambles}.
For both asset classes, we observe an overall positive market sentiment and brief epidemic-like social media activity, leading to increased trade volume activity, causing the creation of multiple risky bubbles \cite{phillips2018mutual, froot1989intrinsic}.
Such epidemic like social media activity and bubble creation exhibit scale-free dynamics since their impact decays as a powerlaw distribution over time \cite{phillips2017predicting}. 
\textsc{mbhn} outperforms its Euclidean variant as it is better able to represent the scale-free dynamics via hyperbolic learning \cite{chen2019sentiment}.
Furthermore, we observe multiple cryptocurrency mentions in a single social media post suggesting that bubble explosivity in one cryptocoin may induce bubbles in another crypto \cite{agosto2020financial}. 
These observations demonstrate the practical applicability of CryptoBubbles quantitative trading as it can scale to forecast multiple risk bubbles.

\paragraph{\textbf{Contextualizing impact of social media hype and meme stocks}}
As shown in Figure \ref{fig:curr}, we note that for meme-stocks like GameStop (GME), Reddit posts consistently display a social media \textit{hype} growing like a chain reaction consecutively creating a bubble \cite{angel2021gamestonk}.
The price value of such meme stocks follows sentiment-driven pricing where more media traffic, more positive tones, more argument leads to higher returns in a short squeeze \cite{chohan2021counter,hu2021rise}. 
We observe that \textsc{mbhn} via hyperbolic learning is able to capture the social media hype of meme-stocks where the intensity of the psychological contagion among retail investors on social media drive asset price fluctuations \cite{semenova2021reddit}. 

\section{Conclusion}
Building on the popularity of speculative social media trading, 
we present CryptoBubbles, a challenging multi-span crypto bubble forecasting task, and dataset. 
We explore the power-law dynamics of social media hype and propose \textsc{mbhn} which learns from the power-law dynamics via hyperbolic learning. 
We curate a Reddit test set to evaluate our models under zero-shot and cold-start settings. 
Through extensive analysis, we show the practical applicability of CryptoBubbles and publicly release CryptoBubbles and our hyperbolic models.

\section*{Acknowledgements}
The research work of Paolo Rosso was partially funded by the Generalitat Valenciana under DeepPattern (PROMETEO/2019/121). 
We would like to thank the Financial Services Innovation Lab at Georgia Institute of Technology for their generous support.

\section{Ethical Considerations}
While data is essential in making models like \textsc{mbhn} effective,
we must work within the purview of acceptable
privacy practices to avoid coercion and intrusive
treatment. To that end, we utilize publicly available
Twitter and Reddit data in a purely observational and non-intrusive manner. 
Although informed consent of each user
was not sought as it may be deemed coercive, 
We follow all ethical regulations set by Twitter and Reddit.

There is an ethical imperative implicit in this growing influence of automation in market behavior, and it is worthy of serious study~\cite{hurlburt2009ethical,cooper2020ethics}.
Since financial markets are transparent~\cite{bloomfield1999market}, and heavily regulated~\cite{edwards1996new}, we discuss the ethical considerations pertaining to our work.
Following ~\cite{cooper2016mysterious}, we emphasize on three ethical criteria for automated trading systems and discuss \textsc{mbhn}'s design with respect to these criteria.

\paragraph{\textbf{Prudent System}} A prudent system \textit{"demands adherence to processes that reliably produce strategies with desirable characteristics such as minimizing risk, and generating revenue in excess of its costs over a period acceptable to its investors"~\cite{longstreth1986modern}}. \textsc{mbhn} is directly optimized towards high risk bubble forecasting.

\paragraph{\textbf{Blocking Price Discovery}} A trading system should not block price discovery and not interfere with the ability of other market participants to add to their own information~\cite{angel2013fairness}. For example, placing an extremely large volume of orders to block competitor's messages \textit{(Quote Stuffing)} or intentionally trading with itself to create the illusion of market activity \textit{(Wash Trading)}. \textsc{mbhn} does not block price discovery in any form.

\paragraph{\textbf{Circumventing Price Discovery}} A trading system should not hide information, such as by participating in dark pools or placing hidden orders~\cite{zhu2014dark}. We evaluate \textsc{mbhn} only on public data in regulated markets.

Despite these considerations, it is possible for \textsc{mbhn}, just as any other automated trading system, to be exploited to hinder market fairness. 
We follow broad ethical guidelines to design \textsc{mbhn} and encourage readers to follow both regulatory and ethical considerations pertaining to the market.

\bibliography{acl}
\bibliographystyle{acl_natbib}

\appendix

\section{Evaluation Metrics}
\subsection{MCC}
Considering imbalance in our dataset we use MCC to to evaluate all our models as it takes into account true and false positives and negatives and is generally regarded as a balanced measure.
The MCC is in essence a correlation coefficient value between -1 and +1. A coefficient of +1 represents a perfect prediction, 0 an average random prediction and -1 an inverse prediction. 
Formally, given a confusion matrix $\begin{pmatrix} tp & fn \\ fp & tn\end{pmatrix}$ MCC is defined as,

\begin{equation}
    \small
    MCC = \frac{(tp\times tn) - (fp\times fn)}{\sqrt{(tp + fp)(tp + fn)(tn + fp)(tn + fn)}}
\end{equation}

\subsection{F1 score}
F1 score is the harmonic mean between precision and recall. The F1 score is defined as, 
\begin{equation}
\small
    F1 = \frac{2}{recall^{-1}+precision^{-1}}
\end{equation}
\begin{equation}
\small
    precision = \frac{tp}{tp+fp}
\end{equation}
\begin{equation}
\small
recall = \frac{tp}{tp+fn}
\end{equation}

\subsection{Exact Match}
EM  measures the percentage of overall predicted bubbles that exactly match the true bubble spans. 
\begin{equation}
    \small
    \text{Exact Match}= \frac{tp+fp}{tp+tn+fp+fn}
\end{equation}

\section{Experimental settings}
We implement \textsc{mbhn} using the Pytorch framework. The Hyperbolic module of \textsc{mbhn} is based on the implementation \footnote{\url{https://github.com/ferrine/hyrnn}}). Our \textsc{mbhn} has a total of 858 and 44,520 parameters with price and text as input respectively.  We utilize the grid search to find all optimal hyperparameters based on the validation MCC scores for all models. As an optimiser we use Adam with \(\beta_1=0.9\) and \(\beta_2=0.999\) and an L2 weight decay of \(1e-5\). We use early stopping based on the Accuracy score over the validation set.

\section{Dataset Curation And Preprocessing}
To create CryptoBubbles dataset, we select top 9 cryptocurrency exchanges and choose around 50 most traded cryptocoins by volume from each exchange and obtain 456 cryptos in total as shown in Table \ref{exchange2coin}. 
For these cryptos we mine 5 years of daily price data consisting of opening, closing, highest and lowest prices from 1\textsuperscript{st} Mar'16 to 7\textsuperscript{th} Apr'21 using CryptoCompare.\footnote{\url{https://www.cryptocompare.com/}} 
Next, we extract cryptocurrency related tweets under Twitter's official license.  
Following \cite{stocknet}, we extract crypto-specific tweets by querying regex ticker symbols, for instance, “$\text{\$DOGE}\backslash$\text{b}” for \textit{DogeCoin}. 
We mine tweets for the same date range as the price data and obtain  2.4 million tweets. 
We observe that the number of tweets increases every year, suggesting the popularity of speculative trading using social media. 

To identify bubbles, we use the PSY model \cite{phillips2015testing} which is a widely used exuberance detection method in financial time-series analysis \cite{cheung2015crypto,harvey2016tests}.
Following \cite{corbet2018datestamping} we feed the closing prices of each cryptocurrency to the PSY model which outputs date spans for each bubble having labels 1 or 0 denoting whether the day is included in the bubble or not. We rank all of our cryptos by the trade volume and delete some of the lower ranked cryptos having no bubble. After this step we obtain with a set of 404 cryptos.

To generate a data sample belonging to a crypto $c_i$, we first pick $T$ lookback dates and $\tau$ lookahead dates. Now, for each lookback day we randomly choose 15 tweets and chronogically append the tweets. For the corresponding lookahead days we pick the labels generated using the PSY model. We repeat this process with a stride of $3$ between the starting lookback dates of previous and next sample. The start (or end) indexes of the bubbles are the index where the bubble started (ended). 

We curate zero-shot Reddit data to test our model's ability to generalize across different asset classes and social media platforms. 
We analyze 12 meme crypto and 17 meme equities (For instance, GameStop, and DOGE) selected based on social media activity over 15 months from 15\textsuperscript{th} Jan'20 to 3\textsuperscript{rd} April'21 as shown in Table \ref{reddit_coins}.
We mine Reddit posts and comments from top trading subreddits such as r/wallstreetbets using the PushshiftAPI.\footnote{\url{https://github.com/pushshift/api}}
We scrape daily price data using Yahoo Finance for equities and CoinGecko for cryptos and use the PSY model \cite{Phillips2015} to identify bubbles. 
We follow the same sample generation process that we use for Twitter data.
We note that zero-shot data establishes a challenging environment for evaluating CryptoBubbles since it contains varied post lengths and unseen asset classes. 

\section{Data Annotation by Financial Analysts}
To further review the ground-truth annotations produced by the PSY model, all annotations were reviewed by five experienced financial analysts achieving a Cohen's $\kappa$ of 0.93.
We recruited these reviewers through an online form, paid them for their work and no ethical considerations were required since we use public financial data. 
We find that the reviewers agree with the annotations for 90\% of the bubbles. During reviewer disagreement (5\% bubbles), we took the majority of all annotators. For 5\% of the bubbles all reviewers agreed that the annotations were incorrect, during which we considered the annotations proposed by the analysts. 

\begin{table}
\small
\centering
\setlength{\tabcolsep}{1.8pt} 

\begin{tabular}{@{}ll@{}}
\toprule
\textbf{Exchange}                  & \textbf{Coin Tickers}               \\ \midrule
\multirow{12}{*}{\textbf{Binance}} & STX, BNB, REN, NKN, COCOS,          \\
                                   & HBAR, ORN, ARDR, TRXUP, BAND,       \\
                                   & WRX, ONE, MTL, EUR, ETHUP,           \\
                                   & VET, KMD, GTO, LTCUP, XRPUP,         \\
                                   & FET, SXP, EOSUP, ERD, UMA, PNT,     \\
                                   & MATIC, ADAUP, BULL, ANKR, RUNE,      \\
                                   & BZRX, LTO, CHR, WIN, MFT, KSM,        \\
                                   & STRAX, VITE, ENJ, NULS, BNBUP,       \\
                                   & KEY, STPT, XTZUP, STMX, SAND,        \\
                                   & BKRW, TROY, ONT, DOTUP, BEL,         \\
                                   & LINKUP, AION, GXS, BEAR, FLM,        \\
                                   & BTCUP, UNIUP, CTSI, DOCK, HIVE,     \\
                                   & DOTDOWN, DUSK, SUSHI, LINKDOWN,     \\
                                   & NPXS, PERL, XTZDOWN, TRXDOWN, \\ 
                                   & DOT, ETHDOWN, WAN, BNBDOWN, \\  
                                   & EOSDOWN, XRPDOWN, ADADOWN, FIO, \\
                                   & RVN, CHZ, LTCDOWN, WING, UNIDOWN, \\
                                   & HOT, BTCDOWN, COTI, BNBBULL, \\
                                   & BNBBEAR \\
                                   \midrule
\multirow{24}{*}{\textbf{Gateio}}  & OCEAN, PI, DDD, FTI, DRGN,           \\
                                   & XRPBEAR, WNXM, YFV, IHT, SNET,       \\
                                   & NBS, MAN, GTC, OCN, SKM, REQ,        \\
                                   & FIL, LBK, CELR, BLZ, SENC, DKA,      \\
                                   & LIEN, KIN, OAX, JNT, SBTC, BCN,      \\
                                   & MDA, EOSBEAR, CDT, OPEN, COFI,       \\
                                   & HNS, RCN, PEARL, NAX, POWR, ETHBEAR, \\
                                   & TFUEL, SWAP, SOP, MBL, FUEL,         \\
                                   & GAS, REM, MXC, CORN, QSP,            \\
                                   & RVC, DX, XVG, QLC, ZPT, PCX,         \\
                                   & HSC, CREAM, OM, AVAX, QKC, ALY,      \\
                                   & YAM, ADEL, AGS, VTHO, ZSC, COMP,     \\
                                   & BEAM, BCDN, BU, MKR, ELEC, PST,      \\
                                   & DOS, XRPBULL, BAT, RFR, DREP,        \\
                                   & GRIN, DPY, DILI, DOGE, BOT, TOMO,    \\
                                   & PAY, SMT, QASH, XMC, NEO, EOSDAC,    \\
                                   & SOUL, GNX, GARD, DCR, YAMV2,         \\
                                   & ZIL, DATA, SALT, KAVA, GSE, MINI,    \\
                                   & USDG, MIX, XEM, RLC, MDT, TNT,       \\
                                   & BCX, NBOT, KLAY, YFI, XTZ,           \\
                                   & JFI, TAI, GOF, CKB, NAS, BNTY,       \\
                                   & MET, IOTX, EOSBULL, BTMX,            \\
                                   & GMAT, LRN, SERO, RED, LEND,         \\
                                   & OIN, RATING, SFG, DBC, SASHIMI     \\
                                   & KAI, AMPL, RSV, MTA, MTV,  \\
                                   & KTON, AR, BTO, PHA, ETHBULL \\
                                    
                                   \midrule
\multirow{17}{*}{\textbf{OKEX}}    & ACT, DIA, CAI, CNTM, ANW, ZEN,       \\
                                   & ABT, WTC, UGC, WBTC, WXT,            \\
                                   & INT, XUC, TMTG, CMT, DMD,            \\
                                   & CVP, QUN, GUSD, RNT, CELO,           \\
                                   & BLOC, XSR, TCT, HPB, YOU, SUN,       \\
                                   & STORJ, CRO, EM, BTT, XAS, WGRT,      \\
                                   & CHAT, DEP, CIC, HC, RIO, APM,        \\
                                   & KNC, LINK, MOF, OKB, UTK, MITH,      \\
                                   & ARK, UBTC, FTM, PPT, XPR, BSV,       \\
                                   & VIB, JST, NDN, BCD, LRC, FSN,        \\
                                   & DGD, MXT, DGB, CVT, LBA, SWRV,       \\
                                   & XPO, BHP, AE, TRADE, CVC, YFII,      \\
                                   & BTG, FRONT, PRA, PLG, SWFTC,         \\
                                   & ZYRO, OF, TRUE, DNA, HDAO,           \\
                                   & BTM, ELF, EGT, YEE, AST, DMG,        \\
                                   & KCASH, ROAD, SNT, ORBS, APIX,        \\
                                   & ALV, TOPC, BNT, AERGO, RFUEL,        \\
                                   & QTUM, SRM, DENT, IQ, DHT \\
                                   \midrule
\end{tabular}
\end{table}
\begin{table}[t]
\small
\centering
\setlength{\tabcolsep}{1.8pt} 
\begin{tabular}{@{}ll@{}}
\toprule
\textbf{Exchange}    & \textbf{Coin Tickers}             \\ \midrule
\textbf{Bitfinex}    & FUN, DTX, OMG, XAUT, ESS, EGLD,    \\
                     & NEC, GOT, ETC, RRB, XMR, EDO,      \\
                     & RRT, LYM, CLO, ANT, XRP, ONL,      \\
                     & POA, RIF, GNT, SAN, UNI, ETP,      \\
                     & SWM, EOS, BTSE, CND, CNN, PNK,     \\
                     & UFR, ZEC, AGI, RINGX, AUC, LEO         \\
                     & TRX, XRA, WAX, ETH, DTH, BOX           \\ \midrule
\textbf{HuobiPro}                  & HPT, NODE, BIX, KAN, IRIS, SOC,      \\
                                   & LAMB, NANO, LOL, LOOM, ELA,          \\
                                   & OGO, AKRO, GT, OGN, TT, WICC,        \\
                                   & GLM, RSR, VSYS, RUFF, VIDY, BHD,     \\
                                   & NEW, HBC, PVT, MX, MDS, FIRO,       \\
                                   & THETA, NEXO, MANA, CRV, HT,         \\
                                   & LET, FOR, ATP, GXC, DTA, STEEM,      \\
                                   & CTXC, FTT, NEST, ARPA, IOST,         \\
                                   & HIT, DF, ITC, MCO, ACH, BTS         \\
                                   & TRB, WAXP, LUNA, RING \\
                                   \midrule
\textbf{Coinbase}    & CGLD, REP, NMR, XLM, LTC, ZRX, BTC     \\ \midrule
\textbf{Binance USA} & ATOM, WAVES, HNT, SOL, ALGO, IOTA \\ \midrule
\textbf{Kraken}      & ICX, ADA, SC, DAI, PAXG, GBP,      \\
                     & BAL, DASH, SNX, MLN, LSK,          \\
                     & GNO, BCH, AUD                          \\ \midrule
\textbf{Gemini}      & AMP, OXT                          \\ \bottomrule
\end{tabular}
\caption{Exchanges with their corresponding crypto tickers.}
\label{exchange2coin}
\end{table}

\begin{table}[t]
\centering
\small
\begin{tabular}{@{}ll@{}}
\toprule
\textbf{Equity} & \textbf{Crypto} \\ \midrule
AMC             & BAN             \\
BB              & BASED           \\
BBBY            & CAT    \\
BLIAQ  & DOGE            \\
CCIV   & DOGEC           \\
EXPR            & HOGE            \\
JAGX            & MONA            \\
KOSS            & PPBLZ           \\
LGND            & SUSHI           \\
NAKD            & TACO            \\
NIO             & YAM             \\
NOK             & GRLC            \\
PLTR            & -               \\
SNDL            & -               \\
THCB            & -               \\
ZOM             & -               \\
GME             & -               \\ \bottomrule
\end{tabular}
\caption{Equity and cryptos used for Reddit Data.}
\label{reddit_coins}
\end{table}

\section{Limitations}
Time-series forecasting is a fundamental frontier in deep learning. Hyperbolic learning and RNNs are ubiquitous frameworks that can encode temporal relationships between entities. 
By advancing existing RNN approaches
and providing flexibility for RNNs to capture different intrinsic features from hyperbolic spaces, \textsc{mbhn} can be used to represent a wide range of complex streams of data for various applications
such as social influence prediction, healthcare and financial applications. 
One potential issue of \textsc{mbhn}, like many other RNN based models, is that it provides limited interpretability to its
outputs. In the future, we will  look into this to enhance the explainability of new
time-series forecasting architectures, making such RNN architectures applicable in more critical applications such as medical domains.

\end{document}